\documentclass[10pt,letterpaper]{article}
\usepackage[top=0.85in,left=2.75in,footskip=0.75in]{geometry}

\usepackage{graphicx}
\usepackage{bmpsize}
\usepackage{array}
\usepackage{multirow}
\usepackage{epstopdf}
\usepackage{listings}
\usepackage{subfig}
\usepackage{url}
\usepackage{amsmath}
\usepackage{amssymb}
\usepackage{csquotes}

\usepackage{amsmath,amssymb}
\usepackage{changepage}
\usepackage[utf8x]{inputenc}
\usepackage{textcomp,marvosym}
\usepackage{cite}
\usepackage{nameref}
\usepackage[right]{lineno}
\usepackage{microtype}
\DisableLigatures[f]{encoding = *, family = * }
\usepackage[table]{xcolor}
\usepackage{array}
\newcolumntype{+}{!{\vrule width 2pt}}

\newlength\savedwidth

\raggedright
\usepackage[aboveskip=1pt,labelfont=bf,labelsep=period,justification=raggedright,singlelinecheck=off]{caption}

\makeatletter
\renewcommand{\@biblabel}[1]{\quad#1.}
\makeatother

\usepackage{lastpage,fancyhdr,graphicx}
\usepackage{epstopdf}
\fancyheadoffset[L]{2.25in}
\fancyfootoffset[L]{2.25in}
\begin{document}
\vspace*{0.2in}

\begin{flushleft}
{\Large
\textbf\newline{K-Metamodes: frequency- and ensemble-based distributed k-modes clustering for security analytics}
}
\newline
\\
Andrey Sapegin, Christoph Meinel
\\
\bigskip
Internet Technologies and Systems, Hasso Plattner Institute, Potsdam, Germany
\\
\bigskip

* andrey@sapegin.org

\end{flushleft}
\section*{Abstract}
Nowadays processing of Big Security Data, such as log messages, is commonly used for intrusion detection purposed. Its heterogeneous nature, as well as combination of numerical and categorical attributes does not allow to apply the existing data mining methods directly on the data without feature preprocessing. Therefore, a rather computationally expensive conversion of categorical attributes into vector space should be utilised for analysis of such data. However, a well-known k-modes algorithm allows to cluster the categorical data directly and avoid conversion into the vector space. The existing implementations of k-modes for Big Data processing are ensemble-based and utilise two-step clustering, where data subsets are first clustered independently, whereas the resulting cluster modes are clustered again in order to calculate \textit{metamodes} valid for all data subsets. In this paper, the novel frequency-based distance function is proposed for the second step of ensemble-based k-modes clustering. Besides this, the existing feature discretisation method from the previous work is utilised in order to adapt k-modes for processing of mixed data sets. The resulting k-metamodes algorithm was tested on two public security data sets and reached higher effectiveness in comparison with the previous work.


\section*{Introduction}

Nowadays outlier detection algorithms are widely used in different areas, such as banking, insurances, medicine and of course security. In the latter area security experts apply outlier detection methods in order to identify intrusions or other types of malicious behaviour, which should deviate from ``normal'' patterns in the data. This intrusion detection scenario has several characteristics that makes application of outlier detection methods especially complicated. First of all, intrusion detection should often be performed in the absence of the ground truth. Indeed, even for security experts it would be hardly possible to guarantee that the collected monitoring data does not contain any traces of the malicious behaviour in it. This implies that only more sophisticated unsupervised outlier detection methods could normally be used for processing of such data. Next, the heterogeneous nature of Big Security Data (that are collected from a variety of devices, operating systems and services) makes it hard to stick to specific set of numerical features or metrics. Rather, each data feed has its own fields/format, which requires either conversion to a common log format \cite{OLF} or application of generic outlier detection methods applicable on data without feature preparation \cite{Sapegin2017, thesis}. Finally, security data, such as log messages or network traffic (e.g., TCP dump or derived features) normally contain textual data (username, action, domain name, protocol, etc.), or also numerical data, but of different measures and distribution types, i.e. bytes transferred and number of connections per hour.  Therefore, most existing data mining algorithms cannot be directly applied on such mixed data.

Rather, most existing methods including classic clustering-based approaches require discretisation of continuous numerical data into categories \cite{Garcia2013, Sapegin2017} and conversion of textual data into vector space model in order to apply distance function on data, which is needed to build clusters and find outliers.

Luckily, textual data of security log messages actually represent a limited number of categories and not a free natural text. So instead of applying techniques such as TF-IDF in order to convert log messages into vector space model, a direct conversion using one-hot encoding --- where each unique field value becomes a separate column/dimension --- is normally performed on the data\footnote{Since each term appears only once in the log line, calculation of term frequencies does not bring any benefits.}. Of course, in the large enterprise networks covering data of, for example, 100,000 employees a username column alone will require 100,000 dimensions in the vector space representation, whereas full vector space can contain 500-600 thousand dimensions/columns. Although special programming language structures and algorithms (such as sparse matrices and spherical k-means clustering) allow processing of high-dimensional vector space models \cite{Sapegin2017,thesis}, such conversion still affects performance and may have higher requirements on RAM usage.

Therefore, intrusion detection for Big Security Data may benefit from utilisation of generic clustering-based outlier detection that does not require data conversion into vector space model. The most well-known algorithm for this task is k-modes, which by default uses hamming distance (``simple matching similarity'') and defines modes (``cluster centers'') as a set of most frequent category for each feature/attribute. Since k-modes was originally proposed by Huang et al in 1998 \cite{Huang1998}, many researchers worked on various improvements of this algorithm. 

In 2004, San et al. elaborated on the problem of non-unique cluster modes in k-modes, which ``makes the algorithm unstable depending on mode selection during the clustering process'' \cite{San2004}.
Authors propose to replace modes with ``cluster representatives'', which represents a fuzzy set with all possible categorical values for each attribute from the cluster, whereas each value is characterised by its relative (to the mode/representative) frequency. San et al. also propose a new distance function, where the distance between object and mode/representative is calculated based on the relative frequencies for each attribute in the object.

The similar idea was proposed in 2005 by He et al., where dissimilarity function also takes into account the relative frequency of the attribute, but only if the most frequent attribute category in the mode is equal to the attribute value in the object \cite{He2005}.

Both San et al. and He et al. demonstrated that the frequency-based distance functions allow to achieve higher clustering accuracy. Following it, Ng et al. in 2007 provided a formal proof for the effectiveness of the k-modes with frequency-based dissimilarity measure, as well as confirmed its guaranteed convergence. Next, Cao et al. in 2011 proposed improved dissimilarity measure based on rough set theory \cite{Cao2012}. This novel dissimilarity measure (1) takes into account ``the distribution of attribute values over the whole universe''\footnote{The distance function is based on the similarity measure that takes into account the number of equivalence classes in the whole data set with respect to the attribute value being compared.}, (2) has higher effectiveness (on selected biological and genetical data sets), (3) eliminates some border cases when object assignment to the mode is undetermined and (4) also guarantees convergence for k-modes. Besides that, authors showed both the effectiveness and efficiency of k-modes on large data sets.

The more detailed review of various modification of k-modes algorithm was provided by Goyal and Aggrawal \cite{KModesReview}. The review paper covers not only the improvement of distance functions, but also related work on initialisation of modes and automatic selection of parameter k.

Finally, when distributed computing become widely used, Visalakshi et al. proposed one more important modification of k-modes algorithm, namely the ensemble-based distributed version of k-modes in 2015 \cite{KarthikeyaniVisalakshi2015a}. Under this approach, data are divided into subsets and these subsets are clustered on the different nodes using k-modes. Next, all modes from all subsets are collected by the master node and undergo one more iteration of k-modes clustering in order to calculate global clusters. Authors claim equal or better performance and cluster quality in comparison with non-distributed k-modes as well as classic distributed k-means algorithm \cite{DKM} (authors utilised label/integer encoding in order to apply k-means on categorical data).

However, in the distributed version of k-modes the second step of the algorithm still requires optimisation. Indeed, after all modes from all subsets are collected on the master node, one needs to calculate distances between (1) pairs of modes (from the first step of the algorithm), as well as (2) modes and ``modes of modes'' (which will be called \textit{metamodes} in this paper). Even if the modes themselves were calculated taking relative frequencies of the attribute values into account, there is no existing distance function that can calculate dissimilarity between two modes/representatives containing relative frequencies for all possible attribute values in the cluster.

In this paper we review the distributed k-modes algorithm and propose a novel distance function for clustering of modes (cluster representatives containing frequencies for all possible categorical values for each attribute from the cluster objects). We utilise new distance function for clustering of modes in the second step of the distributed k-modes algorithm. We also prove that the resulting \textit{metamodes} represent global cluster centers more effectively than in the cases when attribute frequencies are discarded after first step of the algorithm. Besides that we also combine distributed k-metamodes with discretisation of numeric data, which allows to apply this clustering method on numeric or mixed (containing both numerical and categorical data) data sets. The resulting algorithm is compared with Hybrid Outlier Detection from related work \cite{Sapegin2017,thesis} and shows similar effectiveness while avoiding computationally expensive conversion into vector space.

The rest of the paper is organised as follows. In Section \ref{sec:kmodes} we describe existing incremental distributed k-modes algorithm. Next, Section \ref{sec:frequency} proposes the novel frequency-based distance function for clustering of modes/representatives. The effectiveness of the k-metamodes with the new distance functions is evaluated in Section \ref{sec:evaluation}, which also contains comparison with existing Hybrid Outlier Detection algorithm. Finally, Section \ref{sec:conclusion} concludes the paper.

\section{Ensemble-based incremental distributed k-modes}\label{sec:kmodes}

Since k-modes is based on the k-means algorithm, it tries to solve the same optimisation problem. Namely, how to partition a set of objects\footnote{For k-means, the set should contain numeric objects only.} $S = {X_1,X_2,...,X_n}$ into $k$ clusters. Formally, this problem P is described as follows \cite{selim1984}:

\begin{equation} \label{eq:1}
\text{Minimise } P(W,Q) = \sum_{l=1}^{k} \sum_{i=1}^{n} w_{i,l}d(X_i,Q_l)
\end{equation}

subject to

\begin{equation} \label{eq:2}
\sum_{l=1}^k w_{i,l} = 1, 1 \leq i \leq n,
\end{equation}

\begin{equation} \label{eq:3}
w_{i,l} \in {0,1}, 1 \leq i \leq n, 1 \leq l \leq k,
\end{equation}

where Q is a set of modes\footnote{For k-means, Q is a set of cluster centers.}, $W = [w_{i,l}]$ is a partition matrix with size of $n \times k$, and $d(X_i,Q_l)$ is distance function between object and mode.

The original k-modes differs from k-means only in the definition of cluster center (which is replaced with mode) and distance function. Therefore, for both algorithms the problem P can be solved by repeating the following steps until P will not converge to the local minimum \cite{Huang1998}:

\begin{itemize}
\item \textbf{step 1:} Fix Q and solve P through finding optimal W. Here the set of modes is fixed and for each object the best mode is identified through calculating the distance between object and mode.

\item \textbf{step 2:} Fix W and solve P through finding optimal Q. Here the modes are recalculated based on the object reassignments from step 1.
\end{itemize}

Thus, in order to identify clusters with k-modes, a distance function between object and mode should be defined\footnote{The same distance function can be used to find distance between two objects, since it will be a border case where mode will be based on one object only.}. The most basic hamming distance can be formally defined as:

\begin{equation} \label{eq:4}
d(X_i,Q_l) = \sum_{j=1}^m \delta(x_{i,j},q_{l,j})
\end{equation}

where

\begin{equation} \label{eq:5}
\delta(x_{i,j},q_{l,j}) = \begin{cases} 0 \; if \; x_{i,j}=q_{l,j}, \\ 1 \; if \; x_{i,j} \neq q_{l,j}
\end{cases}
\end{equation}

Of course, in case of frequency-based modes/representatives, both mode and distance function should be redefined. According to San et al., the frequency-based mode\footnote{Hereafter we will always use the notion ``mode'' when talking about both cluster modes and cluster representatives.} is defined as follows \cite{San2004}:

\begin{equation} \label{eq:6}
Q_l = \{q_{l,1},q_{l,2},...,q_{l,m}\}
\end{equation}

where 

\begin{equation} \label{eq:7}
q_{l,j} = \{(c_j,f_{c_j}) | c_j \in V_j\}
\end{equation}

where $V_j$ is a set of all possible values of the attribute $j$ among all objects in the cluster $S'_l$ with mode $Q_l$. Let us also define the $q'_{l,j}=c'_j$, where $c'_j$ is the most frequent attribute value in $V_j$, so that $f_{c'_j} \geq f_{c_p,j}$ $\forall p$ $|$ $V_j$ \cite{He2005}.

When modes are calculated based on the attribute frequencies, a frequency-based distance function can be applied to calculate distance between object and mode, formally the formula \ref{eq:5} should be replaced with:

\begin{equation} \label{eq:8}
\delta(x_{i,j},q_{l,j}) = \begin{cases} 1 - f_{c_j}(c_j=q'_{l,j}|S'_l) \; if \; x_{i,j}=q'_{l,j}, \\ 1 \; if \; x_{i,j} \neq q'_{l,j}
\end{cases}
\end{equation}

In the ensemble-based k-modes (proposed by Visalakshi et al. in \cite{KarthikeyaniVisalakshi2015a}) this distance function can be applied at the first step of the ensemble-based clustering, where each data subset is clustered independently at the ensemble member (k-modes algorithm instance). Nevertheless, the next steps of the proposed ensemble-based clustering expects clustering of modes themselves, i.e. applies k-modes on collection of modes from all ensemble members. In this case the existing frequency-based distance function that uses formula \ref{eq:8} cannot be applied in order to calculate the distance between two modes while taking into account attribute frequencies calculated on the first step of the algorithm. The only existing solution supposes discarding previous attribute frequencies from the calculation of the distance and treating modes as usual objects\footnote{In order to convert mode to object, the classical definition of mode by Huang can be used, i.e. the mode can be converted back to $q'$ or the set of the most frequent values for each attribute.}. However, this approach might be less effective in comparison to frequency-based distance function.

Therefore, in this paper we propose a novel frequency-based distance function for clustering of modes at the third step of ensemble-based k-modes algorithm. The proposed distance function is described in details in the section below.

\section{Frequency-based distance function for calculation of metamodes}\label{sec:frequency}

Similarly with distance function from formula \ref{eq:4}, the distance function for clustering of modes should be able to calculate distance between clustering object (mode) and cluster center (metamode). While the mode is already defined with formula \ref{eq:7}, the metamode can be defined as a set of all attribute frequencies from all objects with all modes in the meta-cluster. Formally,

\begin{equation}\label{eq:9}
\text{Metamode } Z_t = \{z_{t,1},z_{t,2},...,z_{t,m}\}
\end{equation}

where

\begin{equation}\label{eq:10}
z_{t,j} = \{(c_j,f_{c_j})|c_j \in V'_j\}
\end{equation}

where $V'_j$ is a set of all possible values of the attribute $j$ among all objects in all clusters $S'$ with all modes $Q'$ with metamode $Z_t$. In order to be able to calculate the frequencies of the metamode attributes, it is needed to keep original counts (and not frequencies) of attribute values in the mode, since the frequencies are not scaled to the cluster size. Formally, we redefine mode so that formula \ref{eq:7} becomes:

\begin{equation} \label{eq:11}
q''_{l,j} = \{(c_j,f'_{c_j}) | c_j \in V_j\}
\end{equation}

where $f'$ is the number of occurrences of $c_j$ as a value of attribute $j$ in the cluster $S'_l$ with mode $Q_l$. Here we note that the distance between object and mode can still be calculated using formula \ref{eq:8}, since it is easy to calculate $f_{c_j}$ from $f'_{c_j}$ and, correspondingly $q_{l,j}$ from $q''_{l,j}$:

\begin{equation} \label{eq:12}
(f_{c_j} | c_j \in V_j) = f'_{c_j} / n', 
\end{equation}

where $n'$ is the number of objects $X$ in the cluster $S'_l$.

Thus, both modes and metamodes become a fuzzy sets containing counts for each possible attribute value for each object in the cluster and meta-cluster, while $V'_j \subset V_j$. This allows us to define a frequency-based distance function to find distance between two modes (or mode and metamode) as sum of Euclidean distances for each attribute:

\begin{equation} \label{eq:13}
d(Q_l,Z_t) = \sum_{i=1}^n \sqrt{\sum_{j=1}^m (q_{l,j}-z_{t,j})^2}
\end{equation}

In order to differentiate it from \ref{eq:8}, we will call it as \textit{meta-frequency-based} distance function in this paper.

Please note that in order to calculate both metamode and distance between mode and metamode, we use both $q_{l,j}$ (fuzzy set of attribute frequencies) and $q''_{l,j}$ (fuzzy set of attribute counts), although $q_{l,j}$ can be calculated from $q''_{l,j}$ on the fly.

The new distance function allows to take into account attribute frequencies in the modes for calculation of distance to the metamode. This approach should be more effective than the case when the mode is converted back to $q'$ in order to discard attribute frequencies and threat mode as an object, which allows to apply distance functions from the previous work \cite{Huang1998, San2004, He2005, Ng2007, Cao2012} mentioned in the introduction and also as formulas \ref{eq:5} and \ref{eq:8}. The effectiveness of the k-metamodes with proposed distance function is evaluated on two data sets as described in the next section.

\section{Evaluation of k-metamodes on public KDD Cup 1999 and UNSW-NB15 network data sets}\label{sec:evaluation}

KDD Cup 1999 data set is the most popular security data set for evaluation of machine learning and data mining algorithms \cite{KDD99CUP}. This dataset contains both numerical and categorical features, which makes it perfect example of data that SIEM and IDS systems need to process. However, this data set is already 20 years old and contains a very high (80\%) attack ratio \cite[Section~4.3.2]{thesis}, which makes only first 400,000 records with attack ratio of 9.8\% suitable for evaluation of unsupervised outlier detection algorithms\footnote{With the high attack ratio, such as 80\%, unsupervised outlier detection tends to learn attacks as ``normal'' and therefore is not suitable for processing of such data.}.

\begin{figure}[htb!]
  \centering
  \includegraphics[angle=-90,width=\linewidth]{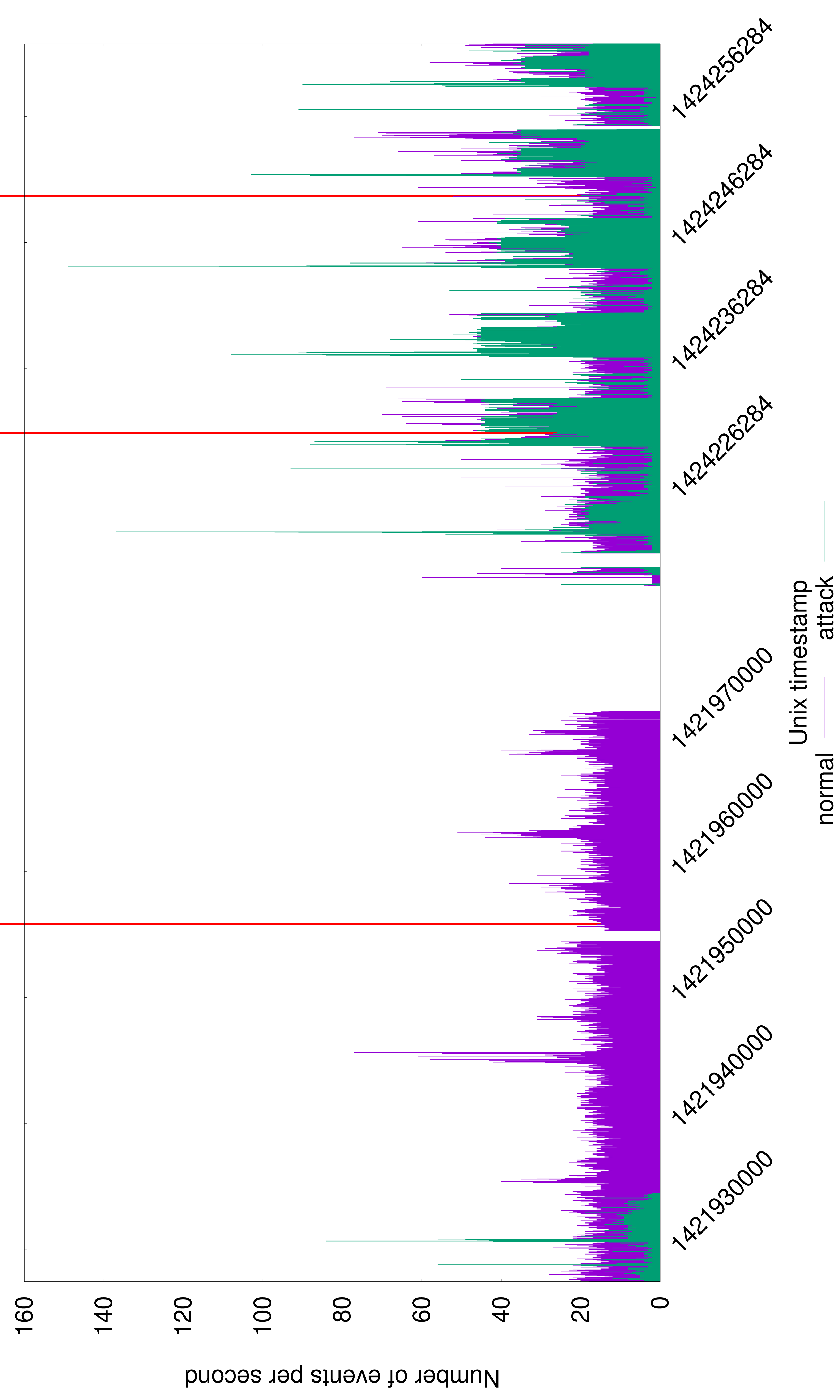}     
  \centering
  \caption{Overview of UNSW-NB15 data set}
  \label{fig:nb15overview}
\end{figure}

In 2015, Moustafa et al. proposed the newer data set that also has lower attack ratio \cite{nb15a,nb15b}. The UNSW-NB15 data set covers two simulation periods: (1) 16 hours on Jan 22, 2015 and (2) 15 hours on Feb 17, 2015. During these periods of time, the raw traffic was collected and later converted with the help of Argus\cite{Argus} and Bro \cite{Bro} into higher-level set of features (similar to KDD Cup 1999 data set) available as CSV files. Although all types of data (raw traffic in pcap format, BRO files, Argus files and CSV) are available for download, we use CSV files for evaluation of k-metamodes. In total, there are 4 CSV files covering both time periods, as shown in Figure \ref{fig:nb15overview}.

Figure \ref{fig:nb15overview} shows rates of normal and attack records per second from both time periods, as well as all 4 CSV files. The simulation periods are recognisable through a gap in the middle of the figure, whereas red lines in the figure show the borders between CSV files. The second time period has much higher attack ratio (10 attacks per second or 25.92\% of data), while first time period is more suitable for evaluation of unsupervised outlier detection algorithms (1 attack per second or 2.09\% of data). Therefore, we only apply k-metamodes on the data from 16 hours of simulation on Jan 22, 2015. In total, this part contains 1,087,203 records, 22,215 of them are related to the malicious activity and listed in Table \ref{tab:attacks}.

\begin{table}[htbp!]
\caption{Number of records per attack category}
\begin{center}
\begin{small}
\begin{tabular}{|>{\centering}c|c<{\centering}|}
\hline
Attack category & Number of records \\ \hline
Analysis & 526 \\ \hline
Backdoors & 534 \\ \hline
DoS & 1167 \\ \hline
Exploits & 5409 \\ \hline
Fuzzers & 5051 \\ \hline
Generic & 7522 \\ \hline
Reconnaissance & 1759 \\ \hline
Shellcode & 223 \\ \hline
Worms & 24 \\ \hline
\end{tabular}
\end{small}
\end{center}
\label{tab:formats}
\end{table}

The k-metamodes outlier detection algorithm was applied on both of described data sets.

In order to apply it on KDD Cup 1999 Data, the data were discretised according to previous work \cite[Section~4.3.2]{thesis}. Due to the fact that k-modes is able to process discretised categorical values only, advanced discretisation --- which produces continuous numerical values in the range between 0 and 1 --- cannot be applied on the data. Therefore, we utilised simple discretisation, sample size of 10,000 records\footnote{This sample size was selected to be the same with \cite[Section~4.3.2]{thesis} in order to be able to compare results with the previous work.} and $k=22$ (number of modes per sample), which is the optimal k value for these data and discretisation type (please see \cite[Section~4.3.6]{thesis} for details). Each sample was selected using random sampling without replacement due to the fact that without random sampling some of KDD Cup data subsets (10,000 records each) will not have enough unique records to initialise 22 modes. Next, frequency-based distance function from Formula \ref{eq:8} was used on both steps of ensemble-based k-modes clustering. As mentioned in the Section \ref{sec:kmodes}, at the second step of the clustering,  attribute frequencies were discarded in order to apply distance function from \ref{eq:8} for calculation of distance between mode and metamodes. Besides that, for the second step of the clustering, we continue using $k'=22$ (number of metamodes). The resulting AUC (Area Under Curve) values for different number of samples are shown in Figure below.

\begin{figure}[htb!]
  \centering
  \includegraphics[angle=-90,width=0.5\linewidth]{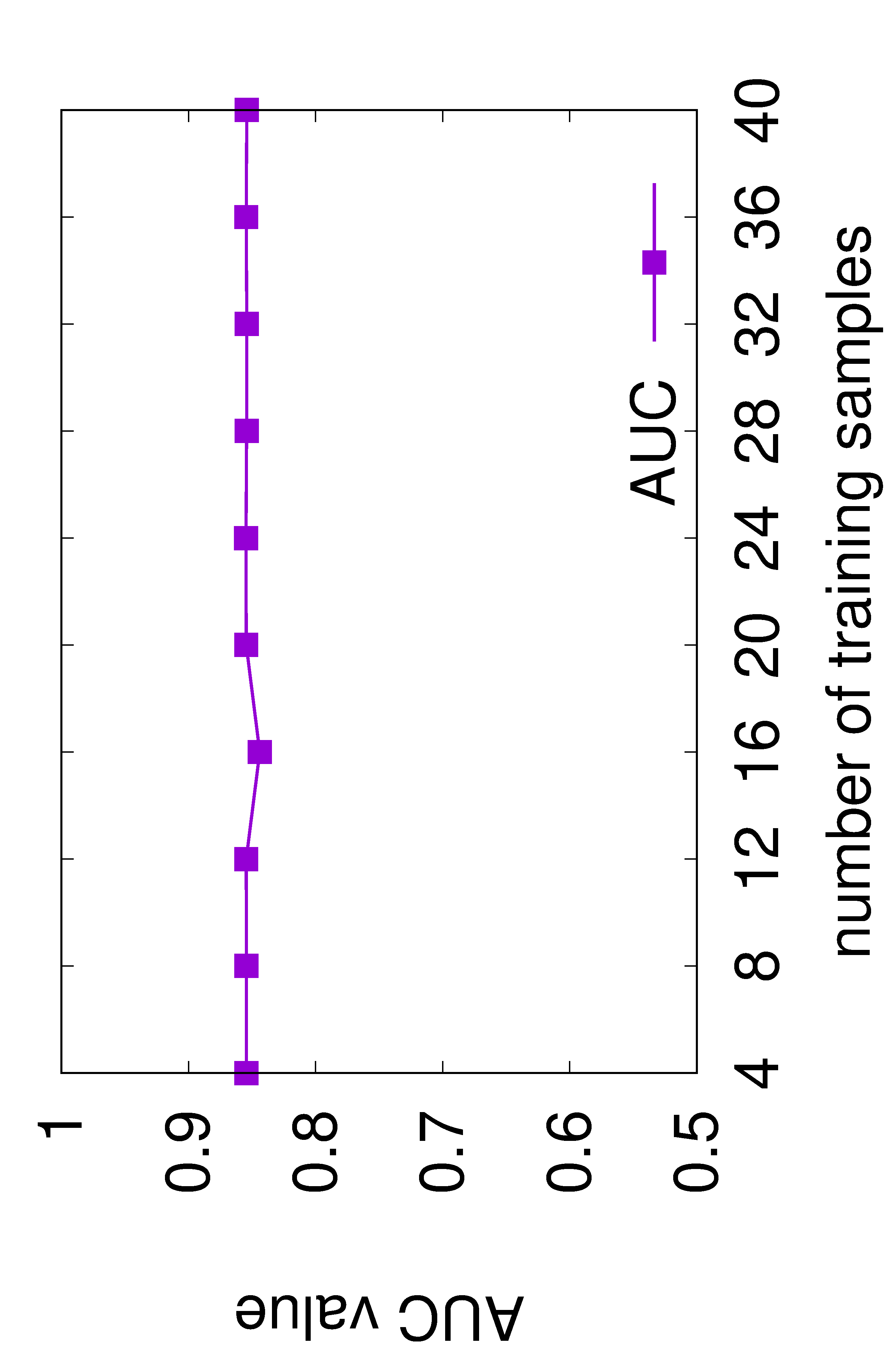}     
  \centering
  \caption{AUC values for KDD Cup 1999 data with the different number of training samples, sample size 10,000, k=22, k'=22, frequency-based distance function, distance to all metamodes as outlier score.}
  \label{fig:KDDCupAUC}
\end{figure}

In Figure \ref{fig:KDDCupAUC}, the AUC value does not show any dependency on the number of training samples, which supports the claim that training on subsamples of data does not decrease the quality of outlier detection \cite[Section~3.2]{subsampling}. Rather, the k-metamodes trained on just 4 samples 10,000 records each (10\% of the data set altogether) allows to achieve the same AUC value as k-metamodes trained on the full data (40 samples).

However, another factor --- namely the measure selected as outlier score --- may affect the effectiveness of the outlier detection. In the previous work, the outliers were clustered together and each outlier from the cluster was assigned the score of the cluster center \cite[Section 4.3.1.1]{thesis}. K-metamodes outlier detection allows to reproduce this approach. Instead of taking distance from each record to all metamodes as outlier score, it is possible to assign outlier score for each record from the corresponding record's mode. Thus, the outlier score turns into the distance from record's mode to all metamodes. With such outlier score, k-metamodes is able to reach \textbf{98,09\% AUC} on the same data\footnote{Calculation of AUC with the proposed outlier score for different number of samples would be unreasonable, since all records from all samples should be clustered in order to assign outlier score to each record}. We provide ROC and Precision-Recall curves in the Figures below.

\begin{figure}[htb!]
  \centering
  \includegraphics[width=0.7\linewidth]{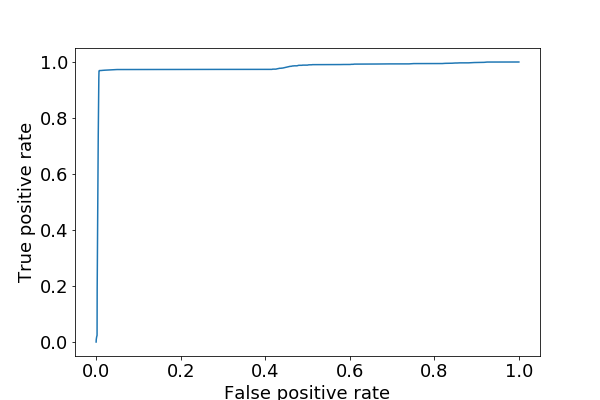}     
  \centering
  \caption{ROC curve for KDD Cup 1999 data with 40 samples, sample size 10,000, k=22, k'=22, frequency-based distance function, distance from record's mode to all metamodes as outlier score.}
  \label{fig:KDDCupFreqROC}
\end{figure}

\begin{figure}[htb!]
  \centering
  \includegraphics[width=0.7\linewidth]{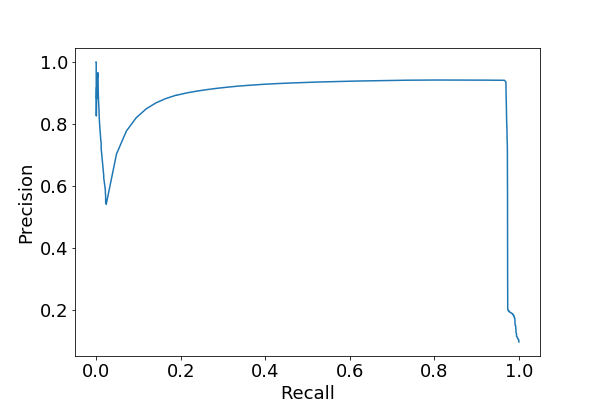}     
  \centering
  \caption{Precision-Recall curve for KDD Cup 1999 data with 40 samples, sample size 10,000, k=22, k'=22, frequency-based distance function, distance from record's mode to all metamodes as outlier score.}
  \label{fig:KDDCupFreqPR}
\end{figure}

Figure \ref{fig:KDDCupFreqROC} shows that the k-metamodes algorithm is able to achieve high true positive rate while still keeping false positive rate low. Moreover, Figure \ref{fig:KDDCupFreqPR} demonstrates that high values for both precision and recall can also be achieved applying k-metamodes on the KDD Cup data.

Next, we apply the same algorithm, but with new distance function proposed in the Section \ref{sec:frequency} on the same data. Since the distance from record's mode to all metamodes as outlier score allows to achieve higher effectiveness, we stick to 40 training samples during the data analysis. Unfortunately, the \textbf{utilisation of proposed distance function allows to reach AUC 97,93\%}, which is slightly worse than original distance function.

\begin{figure}[htb!]
  \centering
  \includegraphics[width=0.7\linewidth]{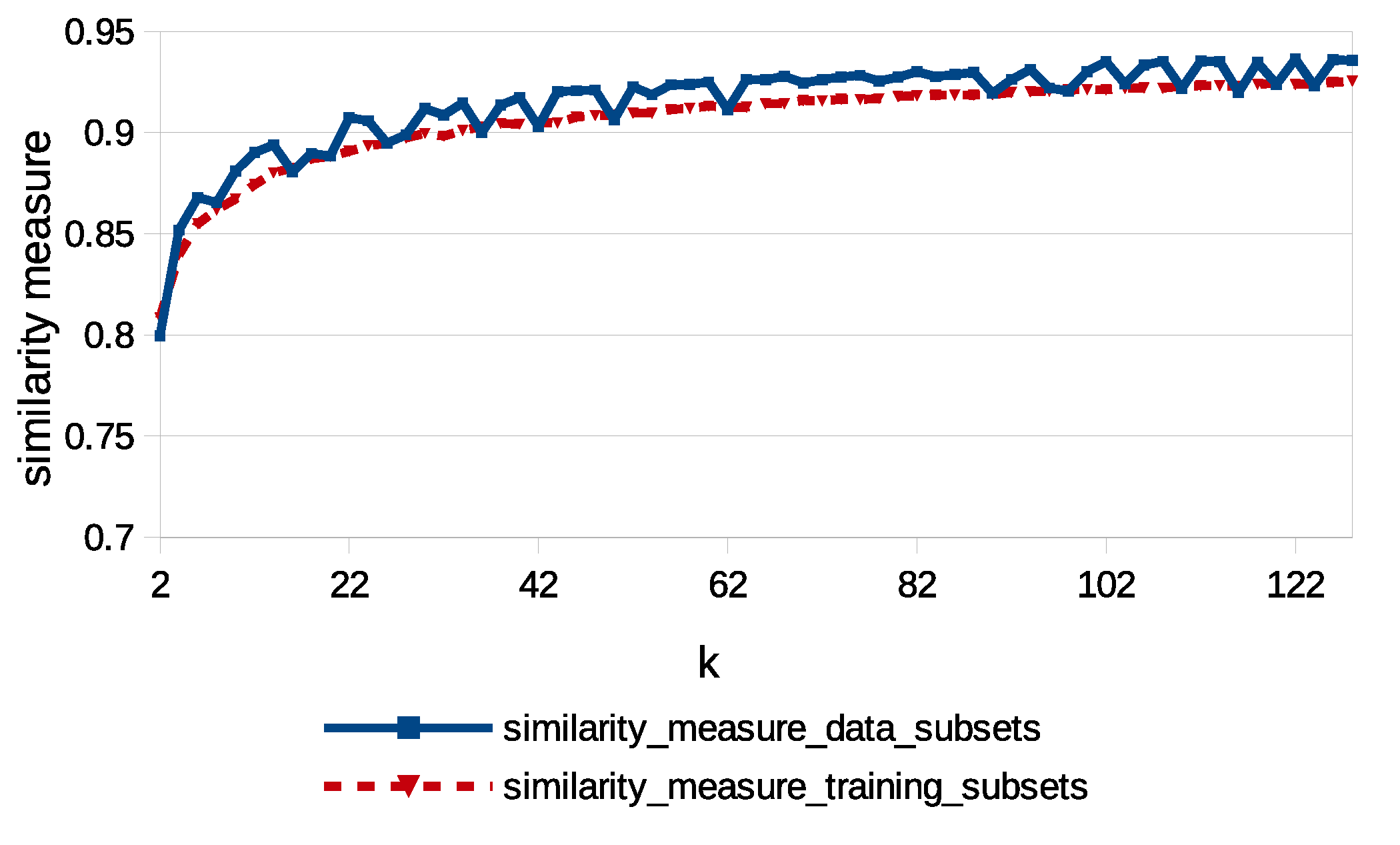}     
  \centering
  \caption{Different number of clusters per sample and corresponding cluster similarity for UNSW-NB15 data set, 50,000 records per sample.}
  \label{fig:NB15optimalK}
\end{figure}

To check if the proposed distance function does not help to achieve higher effectiveness on other data sets as well, we applied k-metamodes on the UNSW-NB15 data. Similarly to KDD Cup data, this data set was discretised in order to convert continuous numerical values into categories. Heuristically, the sample size of 50,000 records was selected. The optimal k was determined according to \cite[Section~4.3.6]{thesis} an equals 36. The Figure \ref{fig:NB15optimalK} provides the charts for different k and corresponding cluster similarity\footnote{Mean cosine similarity between concept vector and each cluster record according to \cite[Section 4.2.4]{thesis}.}.

In Figure \ref{fig:NB15optimalK} the similarity measure is calculated for different k on two types of data subsets. First (called ``data subsets'') are subsets of 50,000 records each, from the UNSW NB15 data set without shuffling. Second (called ``training subsets'') are also subsets of 50,000 records each, but created by applying random sampling without replacement on the UNSW NB15 data set. For both types of subsets, optimal $k$ (equal to 36) allows to reach approximately 90\% cluster similarity.

As we mentioned above, we stick to more effective outlier score, which requires to cluster all records from the dataset with k-metamodes, i.e. the number of ``training'' samples should always cover the data set completely, which implies usage of 22 samples taking into account the data set size of 1,087,203 records and sample size of 50,000 records per sample.

The k-metamodes with frequency-based distance function and parameters discussed above was able to achieve \textbf{AUC of 94,51\%} on this dataset, while k-metamodes with the proposed distance function (defined in the Section \ref{sec:frequency} reaches \textbf{AUC of 96,24\%} on the same dataset. The corresponding Receiver Operating Characteristic, as well as Precision-Recall curves are provided in the Figure below.

 \begin{figure}[htb!]
  \centering
  \subfloat[][ROC curve for UNSW NB15 data with 22 samples, sample size 50,000, k=36, k'=36, frequency-based distance function, distance from record's mode to all metamodes as outlier score.]{
   \includegraphics[width=0.47\linewidth]{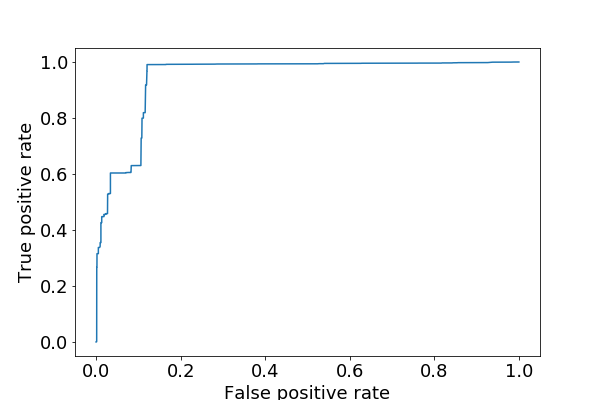}
   \label{fig:NB15_freq_ROC}
  }
  \hspace{0.01\textwidth}
  \subfloat[][ROC curve for UNSW NB15 data with 22 samples, sample size 50,000, k=36, k'=36, meta-frequency-based distance function, distance from record's mode to all metamodes as outlier score.]{
   \includegraphics[width=0.47\linewidth]{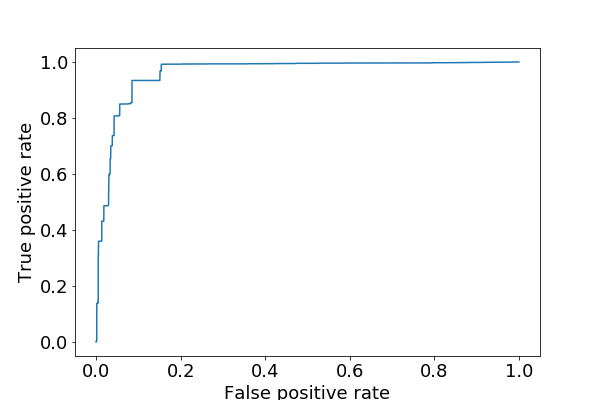}     
   \label{fig:NB15_meta_ROC}
  }
  \hfill
  \subfloat[][Precision-Recall curve for UNSW NB15 data with 22 samples, sample size 50,000, k=36, k'=36, frequency-based distance function, distance from record's mode to all metamodes as outlier score.]{
   \includegraphics[width=0.47\linewidth]{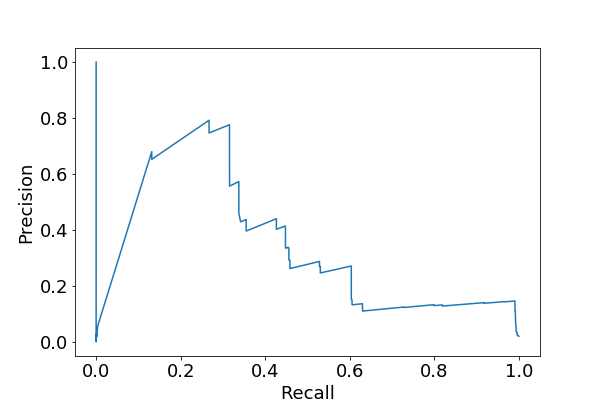}
   \label{fig:NB15_freq_PR}
  }
  \hspace{0.01\textwidth}
  \subfloat[][Precision-Recall curve for UNSW NB15 data with 22 samples, sample size 50,000, k=36, k'=36, meta-frequency-based distance function, distance from record's mode to all metamodes as outlier score.]{
   \includegraphics[width=0.47\linewidth]{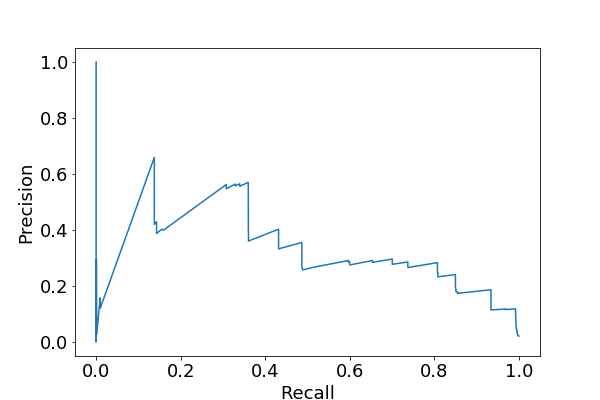}
   \label{fig:NB15_meta_PR}
  }
  \caption{ROC and PR curves for k-metamodes on UNSW NB15 data with both frequency-based (on the left) and meta-frequency-based (on the right) distance functions.}
  \label{fig:NB15ROCPR}
 \end{figure}

In Figure \ref{fig:NB15ROCPR}, both ROC curves for k-metamodes with frequency-based and the proposed meta-frequency-based distance functions show that the algorithm is able to achieve high true positive rate while keeping relatively low false positive rate. In turn, this proves that the k-metamodes is able to place outliers at the top of the algorithm's output by giving higher outlier score to real outliers (true positives). The ROC curve in Figure \ref{fig:NB15_meta_ROC} also shows that with meta-frequency-based distance function, k-metamodes achieves True Positive Rate of more than 90\% while still having False Positive Rate below 10\%. Different to this, without the proposed distance function, True Positive Rate need to be around 60\% in order to keep False Positive Rate below 10\%, as shown in Figure \ref{fig:NB15_freq_ROC}.

On the other hand, both Precision-Recall charts in Figures \ref{fig:NB15_freq_PR} and \ref{fig:NB15_meta_PR} show that k-metamodes applied on the UNSW NB15 data is unable to achieve as high precision, as for KDD Cup data (shown in Figure \ref{fig:KDDCupFreqPR}). For both types of distance functions, the Recall / True Positive Rate of 60\% or more implies relatively low precision which does not even exceed 30\%. However, the precision-recall ratio is not so important for an unsupervised outlier detection algorithm as ROC and AUC, especially in security area. Even if the outlier detection has a high number of false positives, but is able to place true positives at the top of its output, it can still be effectively used for capturing outliers that will be uncaught otherwise (please see \cite[Comparative Evaluation]{Goldstein2016} and \cite[Section~6.1]{thesis} for further details).

We have also checked that increasing the number of clusters per sample, e.g. to $k=100$ does not allow to achieve a higher AUC value. Rather, with the proposed meta-frequency-based distance function and $k=100$ the measured AUC value was 95,13\% (which is less than 96,24\% reached with $k=36$). This fact allows us to conclude that the selected number of clusters per sample has only the minor effect on the outlier detection.

Thus, the results are twofold. On the one hand, the proposed distance function helps to increase AUC from 94,51\% to 96,24\% on the UNSW NB15 data set. On the other hand, on the KDD Cup 1999 data, AUC slightly decreases from 98,09\% to 97,93\%.

In the next section, we compare the k-metamodes with the previous work, namely Hybrid Outlier Detection \cite{Sapegin2017,thesis} to check if k-metamodes is able to achieve higher effectiveness on both data sets.

\subsection{Comparison of k-metamodes with Hybrid Outlier Detection}\label{sec:comparison}

The Hybrid Outlier Detection represents an example of the algorithm that utilises conversion of the features into vector space in order to perform clustering and outlier detection on the categorical data, such as security log messages. Under this algorithm, the data are divided into subsets, whereas each subset undergoes one-hot encoding followed by the clustering using spherical k-means. After the clustering, the concept vectors of clusters are used as training data for the one-class SVM. Each data subset is used to train the corresponding model from the ensemble of one-class SVMs independently. During the application/testing phase, the data is divided into subsets and clustered again in order to check the concept vectors of clusters agains all one-class SVMs from the ensemble. If all models will classify the concept vector as outlier, all the records from the corresponding clusters will be considered outliers and assigned the same outlier score \cite{Sapegin2017,thesis}.

\begin{figure}[htb!]
  \centering
  \includegraphics[angle=-90,width=0.5\linewidth]{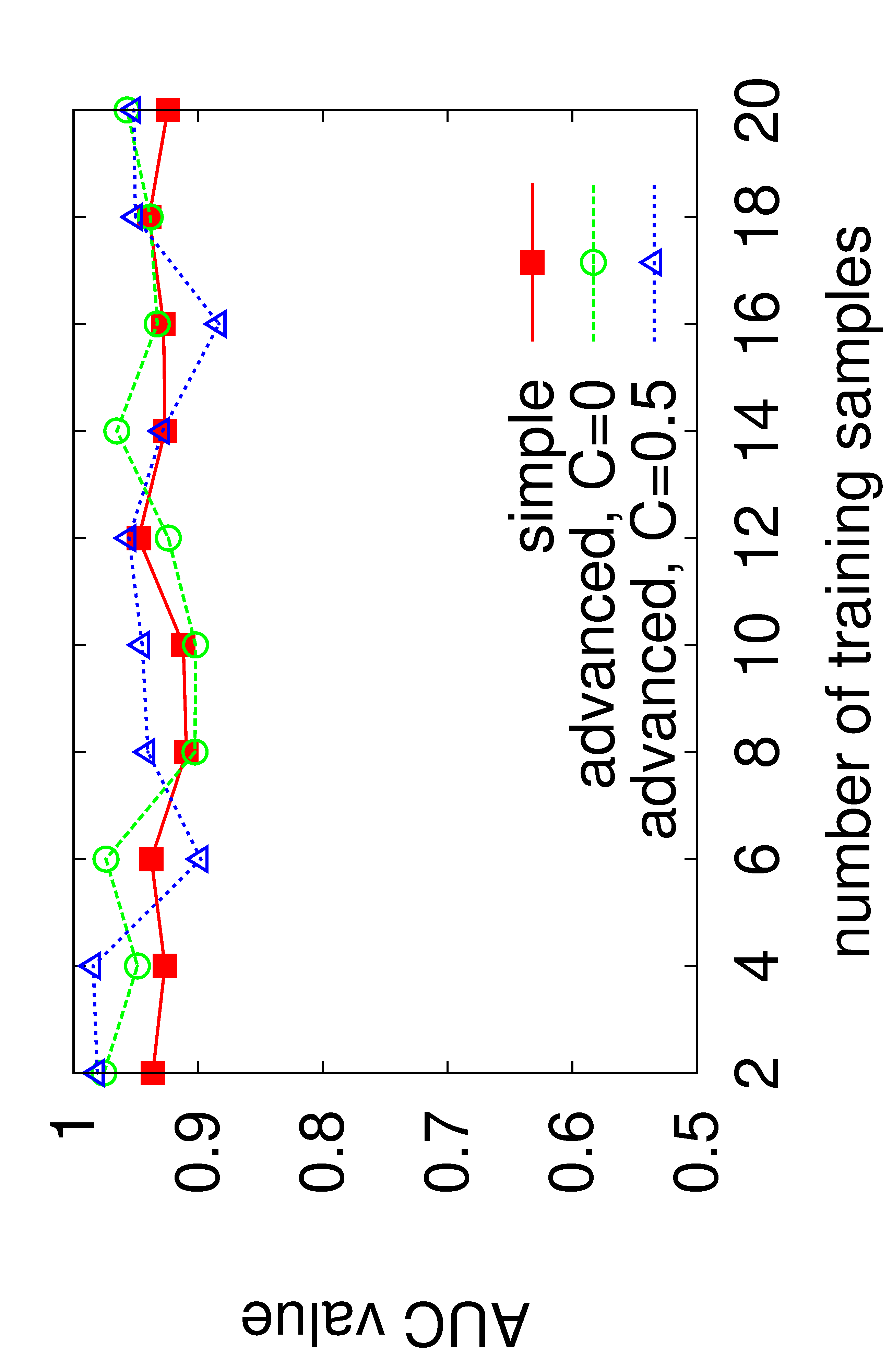}     
  \centering
  \caption{AUC values for Hybrid Outlier Detection on KDD Cup 1999 data with the different number of training samples, sample size 10,000, k=22, simple and advanced (with different coefficient) discretisation; reprinted from \cite{Sapegin2017,thesis}.
  }
  \label{fig:KDDCupAUCHOD}
\end{figure}

In the previous work, Hybrid Outlier Detection was applied on the KDD Cup data set and also achieved a high AUC value, as shown in Figure \ref{fig:KDDCupAUCHOD}.

Figure \ref{fig:KDDCupAUCHOD} provides the measurements of AUC value for different parameters that were used for Hybrid Outlier Detection. The best AUC reached was 98.4\% (with 4 training samples, k = 22 and advanced discretisation with C = 0.5), which is slightly better than the AUC reached by k-metamodes on the same dataset (98,09\%, as mentioned in the previous section). The higher effectiveness of the Hybrid Outlier Detection can be explained by the fact that if k-metamodes is applied on the data sets with mixed data (containing both numerical and categorical features), only simple discretisation might be used to convert numerical features into categorical ones. For Hybrid Outlier Detection (which uses sperical k-means for clustering) it is possible to apply advanced discretisation and retain the difference between original numerical values even though they are discretised into the same category.

\begin{figure}[htb!]
  \centering
  \includegraphics[angle=-90,width=0.5\linewidth]{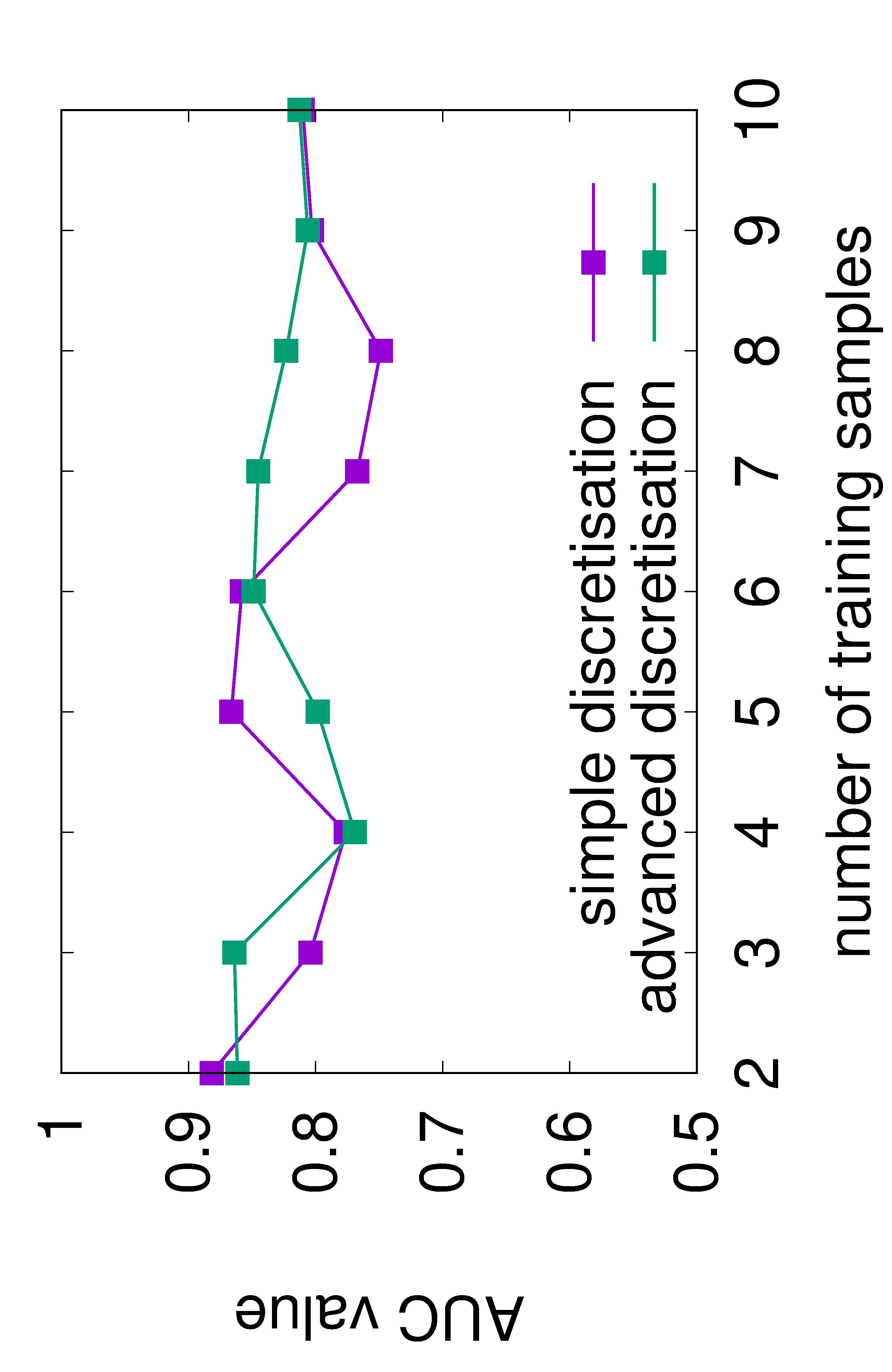}     
  \centering
  \caption{AUC values for Hybrid Outlier Detection on UNSW NB15 data with the different number of training samples, sample size 50,000, k=36, simple and advanced discretisation ($C=0.5$).}
  \label{fig:NB15AUCHOD}
\end{figure}

However, on the UNSW NB15 data set, Hybried Outlier Detection does not outperform k-metamodes, as shown in the Figure \ref{fig:NB15AUCHOD}.

Figure \ref{fig:NB15AUCHOD} presents the AUC values achieved by Hybrid Outlier Detection with different number of training samples\footnote{Similarly to application of Hybrid Outlier Detection on KDD Cup data, for UNSW NB15 data we did not calculate AUC for number of training samples $>$10 (corresponding to approx. half of the dataset), since the algorithm is expected to work best when trained on the subset of data \cite[Section~3.2]{subsampling}.} and different discretisation types ($C=0.5$ was heuristically selected, since this value on of the discretisation coefficient allows the algorithm to achieve the best AUC on the KDD Cup data). Although that the average AUC achieved by HOD was slightly higher for advanced discretisation, the best AUC value was achieved with simple discretisation and equals 88.17\%, which is less than 96,24\% reached by k-metamodes.

Thus, k-metamodes showed almost the same effectiveness on the KDD Cup data as Hybrid Outlier Detection and reached higher AUC value on the UNSW NB15 data set. Next section provides a more details overview on the effectiveness of both algorithms (including both frequency- and meta-frequency-based k-metamodes) and concludes the paper.

\section{Conclusion}\label{sec:conclusion}

All three algorithms --- k-metamodes with frequency-based distance function, k-metamodes with the proposed meta-frequency-based distance function and Hybrid Outlier Detection --- were applied on both KDD Cup 1999 data and UNSW NB15 data and reached high AUC values, which are shown in Figure \ref{fig:AllAUC} below.

\begin{figure}[htb!]
  \centering
  \includegraphics[width=\linewidth]{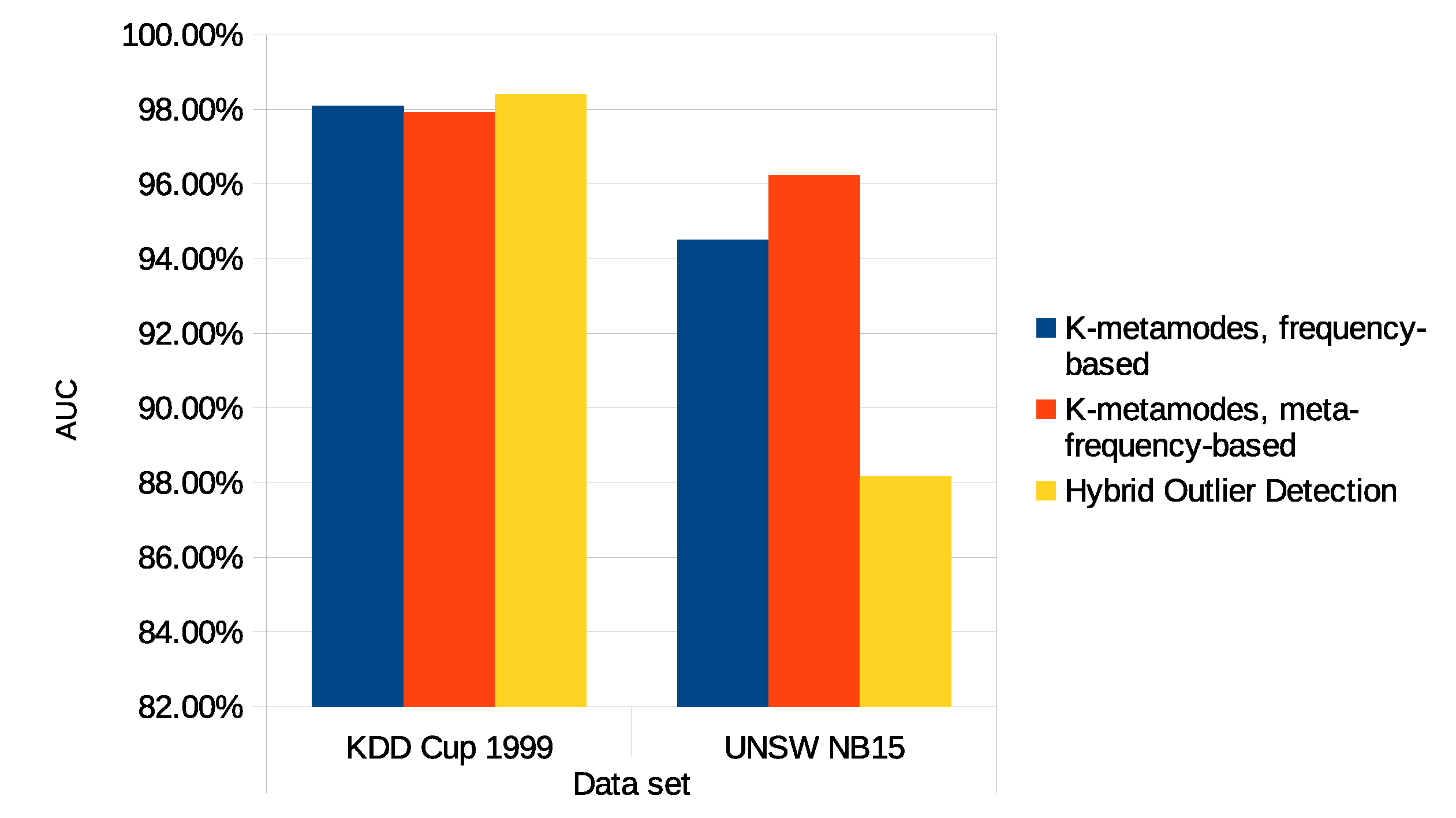}     
  \centering
  \caption{}
  \label{fig:AllAUC}
\end{figure}

Figure \ref{fig:AllAUC} provides an overview of AUC values for all datasets and algorithms tested in this paper. On the KDD Cup 1999 data, all algorithms reach nearly the same AUC value of approximately 98\%. Although the k-metamodes with the proposed meta-frequency-based distance function demonstrates the worst result on these data, the difference with other algorithms might be considered statistically insignificant. Different to this, on the UNSW NB15 data, k-metamodes outperforms Hybrid Outlier Detection, even if advanced feature discretisation was used for the last one. On this dataset, the usage of the proposed meta-frequency-based distance function allows to reach 2\% higher AUC value. Even though 2\% cannot be considered a significant improvement, from ROC curves provided in Figure \ref{fig:NB15ROCPR} we may conclude that the usage of the proposed distance function allows to reach 90\% Recall / True Positive Rate while keeping False Positive Rate as low as 10\%. Without the proposed distance function the corresponding recall will be much lower, i.e. around 60\%.

Thus, in this paper we proposed a novel frequency-based distance function for clustering modes into metamodes within the second step of k-metamodes algorithm. Besides this, we combined k-metamodes with feature discretisation approach from the previous work. The resulting algorithm\footnote{The source code for both k-metamodes with the proposed distance function and Hybrid Outlier Detection is made available on the Github under MIT license \cite{HODgithub,kmetamodesgithub}.} is able to run on mixed data sets containing both numerical and categorical features and allows to reach higher recall for the same False Positive Rate. These improvements are not only relevant for outlier detection in the area of security analytics, where such mixed data sets are rather common, but also for generic Big Data processing cases.

\nolinenumbers

\bibliography{paper}

\end{document}